\documentclass[10pt,twocolumn,letterpaper]{article}

\usepackage{cvpr}
\usepackage{times}
\usepackage{epsfig}
\usepackage{graphicx}
\usepackage{amsmath}
\usepackage{amssymb}
\usepackage{subfigure}

\long\def\symbolfootnote[#1]#2{\begingroup%
\def\thefootnote{\fnsymbol{footnote}}\footnote[#1]{#2}\endgroup}

\newcommand{\denselist}{
\itemsep -2pt\topsep-8pt\partopsep-8pt
}

\newcommand{\BEQAS}{\begin{eqnarray*}}
\newcommand{\EEQAS}{\end{eqnarray*}}

\newcommand{\BIEQAS}{\begin{IEEEeqnarray*}}
\newcommand{\EIEQAS}{\end{IEEEeqnarray*}}

\newcommand{\BEQA}{\begin{eqnarray}}
\newcommand{\EEQA}{\end{eqnarray}}

\newcommand{\BIEQA}{\begin{IEEEeqnarray}}
\newcommand{\EIEQA}{\end{IEEEeqnarray}}

\newcommand{\BITE}{\begin{equation}\left\{ \begin{array}{lll}}
\newcommand{\EITE}{\end{array}\right.\end{equation}}

\newcommand{\BITES}{\begin{equation*}\left\{ \begin{array}{lll}}
\newcommand{\EITES}{\end{array}\right.\end{equation*}}

\newcommand{\eat}[1]{}

\usepackage[pagebackref=true,breaklinks=true,letterpaper=true,colorlinks,bookmarks=false]{hyperref}

\cvprfinalcopy 

\def\cvprPaperID{304} 

\ifcvprfinal\fi
\begin{document}

\title{Detecting events and key actors in multi-person videos}

\author{Vignesh Ramanathan$^{1}$, Jonathan Huang$^{2}$, Sami Abu-El-Haija$^{2}$,
  Alexander Gorban$^{2}$, \\
  Kevin Murphy$^{2}$,
  and Li Fei-Fei$^{1}$\\
  \\
  $^{1}$Stanford University, $^{2}$Google\\
{\tt\small vigneshr@cs.stanford.edu\thanks{This work was done while Vignesh Ramanathan was an intern at Google}
  , jonathanhuang@google.com, haija@google.com,
  gorban@google.com,}\\
  {\tt\small kpmurphy@google.com, feifeili@cs.stanford.edu}
}

\maketitle

\begin{abstract}
Multi-person event recognition is a challenging task, often with many people active in the scene but only a small subset contributing to an actual event. In this paper, we propose a model which learns to detect events in such videos while automatically ``attending'' to the people responsible for the event.  Our model does not use explicit annotations regarding who or where those people are during training and testing. In particular, we track people in videos and use a recurrent neural network (RNN) to represent the track features.  We learn time-varying attention weights to combine these features at each time-instant. The attended features are then processed using another RNN for event detection/classification.  Since most video datasets with multiple people are restricted to a small number of videos, we also collected a new basketball dataset comprising 257 basketball games with 14K event annotations corresponding to 11 event classes.  Our model outperforms state-of-the-art methods for both event classification  and  detection on this new dataset. Additionally, we show that the attention mechanism is able to consistently localize the relevant players.
\end{abstract}

\section{Introduction}

Event recognition and detection in videos has hugely benefited from the
introduction of recent large-scale datasets
\cite{THUMOS,UCF101,Karpathy_CVPR14,MED11,ActivityNet} and models.  However, this is mainly
confined to the domain of single-person actions where the videos contain one
actor performing a primary activity.  Another equally important problem is
event recognition in videos with multiple people. In our work, we present a new
model and dataset for this specific setting.

\begin{figure}[ht!]
  \vspace{-4mm}
\begin{center}
  \includegraphics[width=3.2 in]{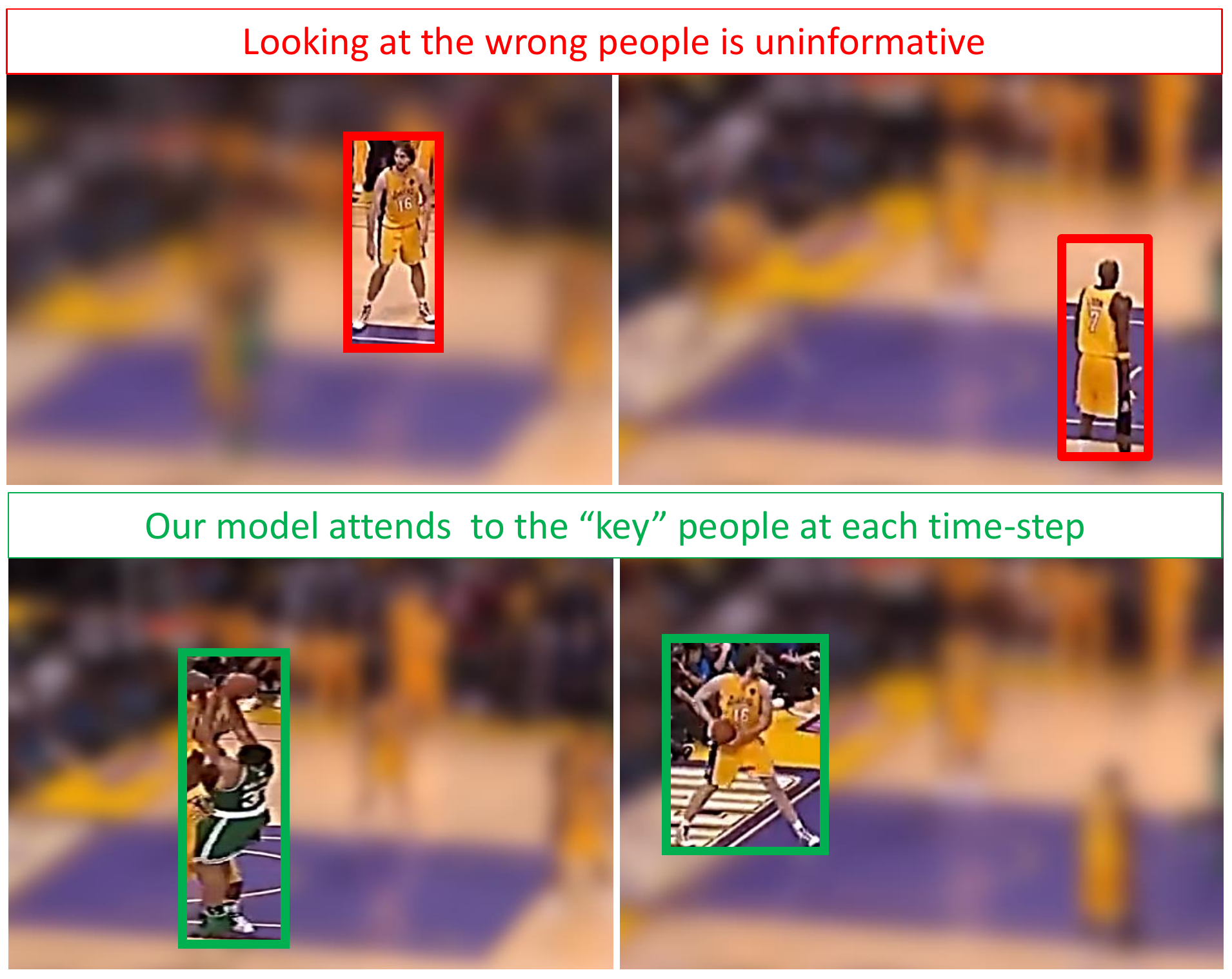}
\end{center}
\caption{Looking at the wrong people in a multi-person event can be very uninformative
  as seen in the basketball video in the first row. However, by observing the correct
  people in the \emph{same video}, we can easily identify the event as a ``2-pointer success" based
  on the shooter and the player throwing the ball into play. We use the same intuition
to recognize the key players for event recognition.}
\vspace{-4mm}
\label{fig:pull_figure}
\end{figure}

Videos captured in sports arenas, market places or other outdoor areas
typically contain multiple people interacting with each other.
Most people are doing ``something'', but not all of them are involved in the main event.
The main event is dominated by a smaller subset of people. For instance, a ``shot" in a game
is determined by one or two people (see Figure~\ref{fig:pull_figure}).
In addition to recognizing the event, it is also important
to isolate these key actors. This is a significant challenge which
differentiates multi-person videos from single-person videos.

Identifying the people responsible for an event is thus an interesting task in its
own right.  However acquiring such annotations is expensive and it is therefore
desirable to use models that do not require
annotations for identifying these key actors during training. This can
also be viewed as a problem of weakly supervised key person identification. In
this paper, we propose a method  to classify events by using a model that is
able to ``attend'' to this subset of key actors.  We  do this without ever
explicitly telling the model who or where the key actors are.

Recently, several papers have proposed to use ``attention" models for aligning
elements from a fixed input to a fixed output.  For example,
\cite{Bahdnau_arxiv14} translate sentences in one language to another language,
attending to different words in the input; \cite{Xu_arxiv15} generate an image-caption,
attending to different regions in the image; and
\cite{Yao_arxiv15} generate a video-caption, attending to different
frames within the video.

\eat{In these settings, the input sequence remains fixed
at all times and the model chooses from this fixed input at each instant.}

In our work, we use attention to decide which of several people is most
relevant to the action being performed; this attention mask can change over
time. Thus we are combining spatial and temporal attention.  Note that while
the person detections vary from one frame to another, they can be associated
across frames through tracking. We show how to use a recurrent neural
network (RNN) to represent information from each track;
the attention model is tasked with selecting the most relevant
track in each frame. In addition to being able to isolate the key actors,
we show that our attention model results in better event recognition.

In order to evaluate our method, we need a large number of videos illustrating
events involving multiple people. Most prior activity and event
recognition datasets focus on actions involving just one or two people.
Multi-person datasets like \cite{Ryoo_ICCV09,VIRAT,Choi_ICCV09} are usually restricted to fewer videos.
Therefore we collected our own dataset.
In particular we propose a new dataset of basketball events with time-stamp annotations for
all occurrences of $11$ different events across $257$ videos each $1.5$ hours
long in length.  This dataset is comparable to the THUMOS \cite{THUMOS}
detection dataset in terms of number of annotations, but contains longer videos
in a multi-person setting.

In summary, the contributions of our paper are as follows.  First, we
introduce a new  large-scale basketball event dataset with 14K dense temporal
annotations for long video sequences.  Second, we show that our method
outperforms state-of-the-art methods for the standard tasks of classifying
isolated clips and of temporally localizing events within longer, untrimmed
videos.  Third, we show that our method learns to attend to the relevant
players, despite never being told which players are relevant in the training
set.

\begin{figure*}[ht!]
  \includegraphics[width=6.5 in]{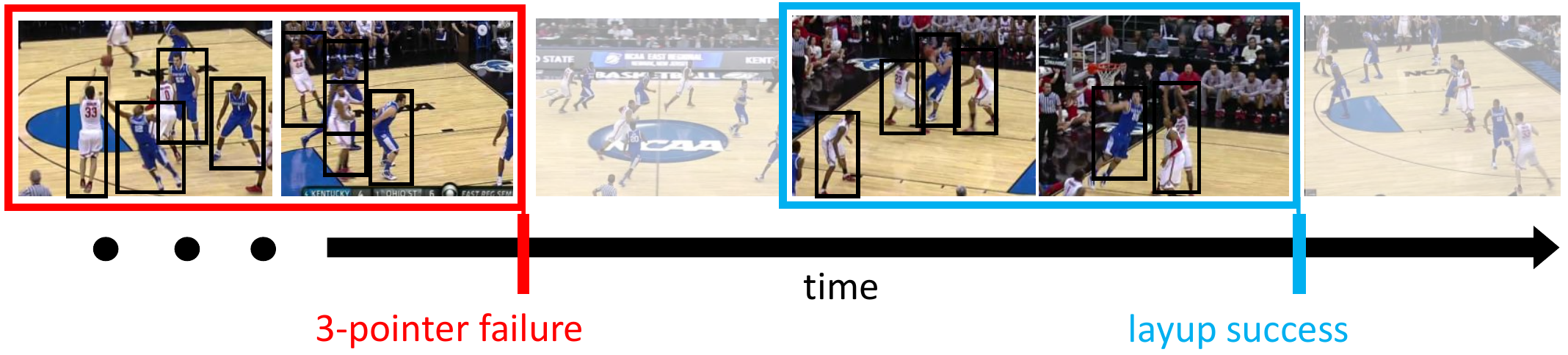}
  \caption{We densely annotate every instance of 11 different basketball events in long basketball
  videos. As shown here, we collected both event time-stamps and an event labels through an
AMT task.}
\end{figure*}

\section{Related Work}

\noindent \textbf{Action recognition in videos}
Traditionally, well engineered features have proved quite effective for video
classification and retrieval tasks
\cite{Dalal_ECCV06,Jain_CVPR13,Jiang_ECCV12,Laptev_CVPR08,
Niebels_ECCV10,Oh_MVA14,Laxton_CVPR07,Oneata_ICCV13,Peng_ECCV14,Sadanand_CVPR12,Schuldt_ICPR04,Wang_BMVC09,Wang_CVPR11}.
The improved dense trajectory (IDT) features \cite{Wang_CVPR11} achieve
competitive results on standard video datasets.  In the last few years,
end-to-end trained deep network models
\cite{Ji_PAMI13,Karpathy_CVPR14,Simonyan_2014,Simonyan_NIPS14,Tran_arxiv14} were shown to be comparable and
at times better than these features for various video tasks.  Other works like
\cite{Wang_arxiv15,Xu_2015,Zha_2015} explore methods for pooling such
features for better performance. Recent works using RNN have achieved
state-of-the-art results for both event recognition and caption-generation
tasks \cite{Donahue_arxiv14,Ng_arxiv15,Srivastava_2015,Yao_arxiv15}.
We follow this line of work with the addition of an attention mechanism
to attend to the event participants.

Another related line of work jointly identifies the region of interest in a video
while recognizing the action.
Gkioxari et al.  \cite{Gkioxari_arxiv14} and Raptis et al. \cite{Raptis_CVPR12}
automatically localize a spatio-temporal tube in a video.
Jain et al. \cite{Jain_CVPR14} merge super-voxels for action localization.
While these methods perform weakly-supervised action localization, they target
single actor videos in short clips where the action is centered around the actor.
Other methods like \cite{Lan_ICCV11,Prest_PAMI13,Tian_CVPR13,Wang_ECCV14} require annotations
during training to localize the action.

\eat{Tian et al. \cite{Tian_CVPR13} train a supervised object detector
while Wang et al. \cite{Wang_ECCV14} train a pose detector with joint annotations
along with the action recognition model.}

\noindent \textbf{Muti-person video analysis}
Activity recognition models for events with well defined group structures such
as parades have been presented in
\cite{Vaswani_CVPR03,Intille_CVIU01,Moore_AAAI02,Khan_ACM05}.  They utilize the
structured layout of participants to identify group events. More
recently, \cite{Lan_PAMI12,Choi_ICCV09,Khodabandeh_arxiv15} use context as a
cue for recognizing interaction-based group activities.  While they work with 
multi-person events, these methods are restricted to smaller
datasets such as UT-Interaction\cite{Ryoo_10}, Collective activity
\cite{Choi_ICCV09} and Nursing home\cite{Lan_PAMI12}.

\noindent \textbf{Attention models}
Itti et al. \cite{Itti_PAMI98} explored the idea of saliency-based attention in
images, with other works like \cite{Shapovalova_NIPS13} using eye-gaze data as
a means for learning attention.  Mnih et al. \cite{Mnih_NIPS14} attend to
regions of varying resolutions in an image through a RNN framework.  Along
similar lines, attention has been used for image classification
\cite{Cao_ICCV15,Gregor_arxiv15,Xiao_arxiv14} and detection
\cite{Ba_arxiv14,Caicedo_ICCV15,Yoo_arxiv15} as well.

Bahdanau et al. \cite{Bahdnau_arxiv14} showed that attention-based RNN models
can effectively align input words to output words for machine translation.
Following this, Xu et al. \cite{Xu_arxiv15} and Yao et al. \cite{Yao_arxiv15}
used attention for image-captioning and video-captioning respectively.
\eat{Following this work, attention was used for aligning regions in an image to
output words for image-captioning \cite{Xu_arxiv15} and frames in a video with
output words for video-captioning \cite{Yao_arxiv15}.} In all these methods,
attention aligns a sequence of input features with words of an output sentence.
However, in our work we use attention to identify the most relevant person to
the overall event during different phases of the event.\eat{We also provide a
better BLSTM based representation for the attended people as
discussed in Section~\ref{sec:methods}.}

\noindent \textbf{Action recognition datasets}
Action recognition in videos has evolved with the introduction of more
sophisticated datasets starting from smaller KTH \cite{Schuldt_ICPR04}, HMDB \cite{HMDB}
to larger , UCF101 \cite{UCF101}, TRECVID-MED \cite{MED11} and Sports-1M
\cite{Karpathy_CVPR14} datasets.  More recently, THUMOS \cite{THUMOS} and
ActivityNet \cite{ActivityNet} also provide a detection setting with temporal
annotations for actions in untrimmed videos.  There are also fine-grained
datasets in specific domains such as MPII cooking \cite{Finegrained_cooking}
and breakfast \cite{Breakfast}.  However, most of these datasets focus on
single-person activities with hardly any need for recognizing the people
responsible for the event. On the other hand, publicly available multi-person
activity datasets like \cite{Ryoo_10,Choi_ICCV09,VIRAT} are restricted to a very small
number of videos.  One of the contributions of our work is a multi-player
basketball dataset with dense temporal event annotations in long videos.

\noindent \textbf{Person detection and tracking}. There is a very
large literature on person detection and tracking.
There are also specific methods for tracking players in
sports videos \cite{Shitrit_ICCV11}.
Here we just mention a few key methods.
For person detection, we use the CNN-based multibox detector from
\cite{Szegedy13}. For person tracking, we use the KLT tracker from
\cite{Veenman_PAMI2001}.
There is also work on player identification (e.g., \cite{Lu2013}), but
in this work, we do not attempt to distinguish players.

\begin{table}[ht!]
\begin{center}
\small
 \begin{tabular}{|l|c|c|}
  \hline
  Event          & \# Videos Train (Test) & Avg. \# people \\ \hline \hline
  3-point succ.    & 895 (188) &  8.35 \\ 
  3-point fail.    & 1934 (401) &  8.42 \\ 
  free-throw succ. & 552 (94) &  7.21\\ 
  free-throw fail. & 344 (41) &  7.85\\  
  layup succ.      & 1212 (233) &  6.89\\ 
  layup fail.      & 1286 (254) &  6.97 \\ 
  2-point succ.    & 1039 (148) &  7.74 \\ 
  2-point fail.    & 2014 (421) &  7.97\\ 
  slam dunk succ.  & 286 (54) &  6.59 \\ 
  slam dunk fail.  & 47 (5) &  6.35\\ 
  steal & 1827 (417) & 7.05\\ \hline  
  \end{tabular}
\end{center}
  \caption{The number of videos per event in our dataset along with
  the average number of people per video corresponding to each of the
events. The number of people is higher than existing datasets for
multi-person event recognition.}
  \label{tab:data_dist}
\end{table}

\section{NCAA Basketball Dataset}
\label{sec:data}

\eat{
Most recent activity detection datasets like THUMOS \cite{THUMOS},
ActivityNet \cite{ActivityNet}, UCF101 
\cite{UCF101}, finegrained cooking \cite{Finegrained_cooking},
contain videos with a single actor performing one activity.
By contrast, we are interested in settings where there are multiple
people in each frame, only a small number of whom are involved with
the  activity. In such a setting, we are interested in recognizing the event
as well as the key participants. A natural choice for such data is
multi-player team sports.
}

A natural choice for collecting multi-person action videos is team sports.
In this paper, we focus on basketball games, although our techniques
are general purpose.
In particular,  we use a subset of the $296$ NCAA games available from 
YouTube.\footnote{https://www.youtube.com/user/ncaaondemand}  These games are
played in different venues over different periods of time.
We only consider the most recent $257$ games, since older games used
slightly different rules than modern basketball.
The videos are typically $1.5$ hours long.
We manually identified $11$ key event types
listed in Tab.~\ref{tab:data_dist}.
In particular, we considered 
5 types of shots, each of which could be successful or failed,
plus a steal event. 

Next we launched an Amazon Mechanical Turk task, where the
annotators were asked to annotate the ``end-point" of these events if and when
they occur in the videos; end-points are usually well-defined (e.g.,
the ball leaves the shooter's hands and lands somewhere else, such as
in the basket).
To determine the starting time, we assumed that each event was 4
seconds long, since it is hard to get raters to agree on when an event
started. 
This gives us enough temporal context to classify each event, while
still being fairly well localized in time.

The videos were randomly split into $212$ training, $12$ validation and $33$
test videos. 
We split each of these videos into 4 second clips (using the
annotation boundaries), and subsampled these to 6fps.
We filter out clips which are not profile shots (such as those shown in
Figure~\ref{fig:model}) using a separately trained classifier; this excludes close-up shots of players,  as
well as shots of the viewers and instant replays.
This resulted in a total of $11436$ training, $856$ validation
and $2256$ test clips, each of which has one of 11 labels.
Note that this is comparable in size to the THUMOS'15 detection
challenge (150 trimmed training instances for each of the $20$ classes and $6553$
untrimmed validation instances). The distribution of annotations across all the
different events is shown in Tab.~\ref{tab:data_dist}. To the best of our
knowledge, this is the first dataset with dense temporal annotations for such
long video sequences.

In addition to annotating the event label and start/end time, we collected AMT
annotations on $850$ video clips in the test set, where the annotators were
asked to mark the position of the ball on the frame where the shooter attempts
a shot.

We also used AMT to annotate the bounding boxes of all the players in a
subset of 9000 frames from the training videos.
We then trained a Multibox detector \cite{Szegedy_arxiv14}
with these annotations, and ran the trained detector on all the videos in our dataset.
We retained all detections above a confidence of 0.5 per frame;
this resulted in 6--8 person detections per clip, as listed in Tab.~\ref{tab:data_dist}.
The multibox model achieves an average overlap of $0.7$ at a recall of $0.8$
with ground-truth bounding boxes in the validation videos.

We plan to release our annotated data, including time stamps, ball
location, and player bounding boxes. 

\section{Our Method}
\label{sec:methods}

All events in a team sport are performed in the same scene by the same set of
players. The only basis for differentiating these events is the action
performed by a small subset of people at a given time.  For instance, a
``steal" event in basketball is completely defined by the action of the player
attempting to pass the ball and the player stealing from him.  To understand
such an event, it is sufficient to observe only the players participating in
the event.

\begin{figure}[t!]
\begin{center}
    \includegraphics[width=3 in]{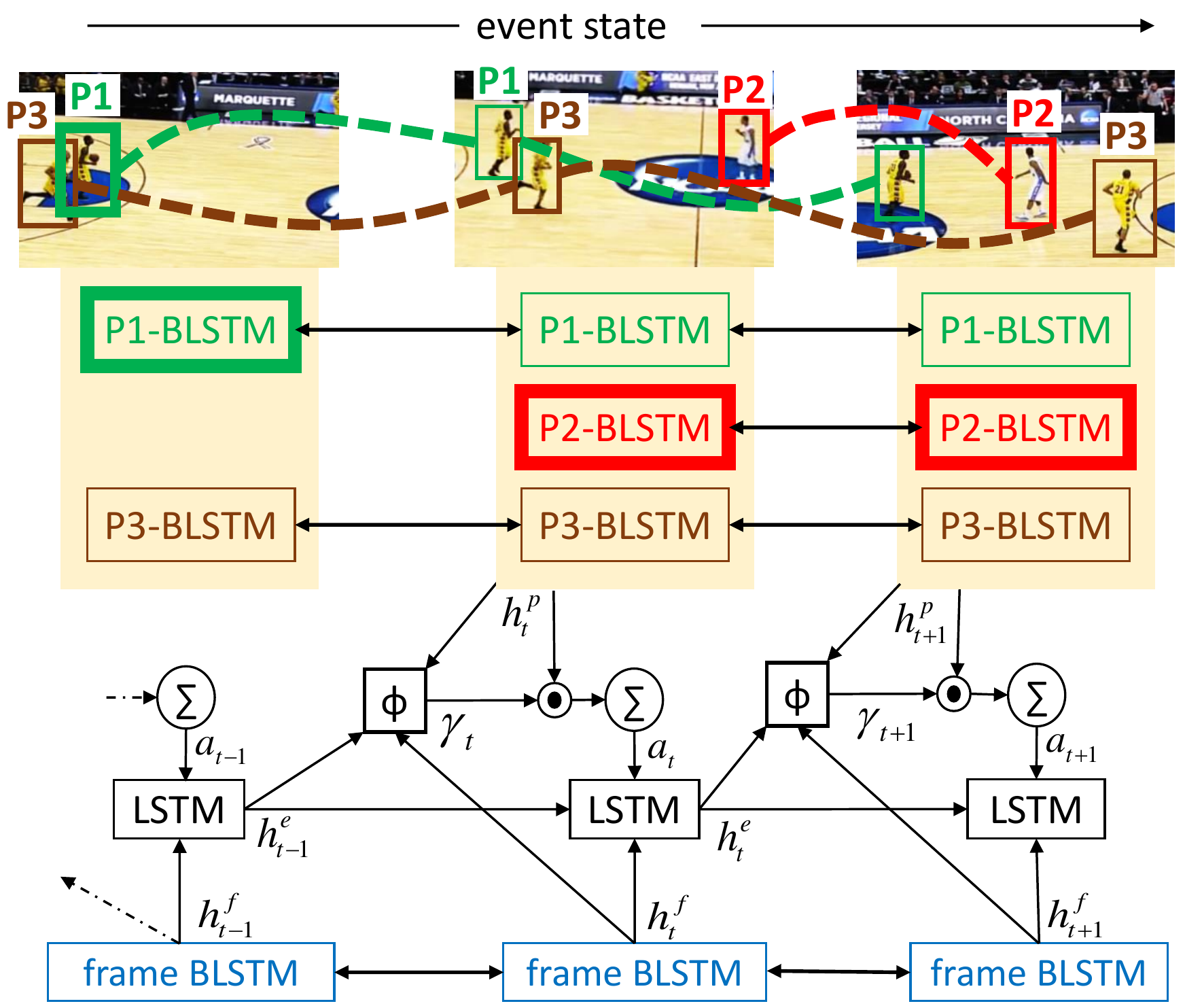}
\end{center}
   \caption{Our model, where each player track is first processed by the
     corresponding BLSTM network (shown in different colors).  $Pi$-BLSTM
     corresponds to the $i$'th player.  The BLSTM hidden-states are then used
     by an attention model to identify the ``key" player at each instant.  The
     thickness of the BLSTM boxes shows the attention weights, and the attended
     person can change with time.  The variables in the model are explained in
     the methods section.  BLSTM stands for ``bidirectional long short term
   memory''. }
\label{fig:model}
\end{figure}

This motivates us to build a model (overview in Fig.~\ref{fig:model})
which can reason about an event by focusing
on specific people during the different phases of the event.
In this section, we describe our unified model for classifying events
and simultaneously identifying the key players.

\subsection{Feature extraction}
\label{sec:feature_extraction}
Each video-frame is represented by a $1024$ dimensional feature vector $f_t$, which is the
activation of the last fully connected layer of the Inception7 network
\cite{Ioffe_arxiv15,Inception7}.  In addition, we compute spatially localized
features for each person in the frame. In particular, we compute a $2805$ dimensional feature
vector $p_{ti}$ which contains both appearance ($1365$ dimensional) and spatial information ($1440$ dimensional) for the
$i$'th player bounding box in frame $t$.
Similar to the RCNN object
detector\cite{Girshick_CVPR14}, the appearance features were extracted by feeding the cropped
and resized player region from the frame through the Inception7 network and
spatially pooling the response from a lower layer. The spatial feature
corresponds to a $32\times 32$ spatial histogram, combined with a spatial pyramid, to
indicate the bounding box location at multiple scales.
While we have only used static CNN representations in our
work, these features can also be easily extended with flow information as
suggested in \cite{Simonyan_NIPS14}.

\subsection{Event classification}

Given $f_t$ and $p_{ti}$ for each frame $t$, our goal
is to train the model to classify the clip into one of 11 categories. As a side
effect of the way we construct our model, we will also be able to identify the
key player in each frame.

First we compute a global context feature for each frame, $h_t^f$, derived from
a bidirectional LSTM applied to the frame-level feature as shown 
by the blue boxes in Fig.~\ref{fig:model}.
This is a concatenation of the hidden states from the forward and reverse LSTM
components of a BLSTM and can be compactly represented as:
\begin{equation}
  h_t^f = \mbox{BLSTM}_{frame}(h_{t-1}^f, h_{t+1}^f, f_t).
\end{equation}Please refer to Graves et al. \cite{Graves_2013}.

Next we use  a unidirectional LSTM to represent the state of the
event at time $t$:
\begin{equation}
  \label{eq:event_lstm}
h_t^e = \mbox{LSTM}(h_{t-1}^e, h_t^f, a_t),
\end{equation}
where $a_t$ is a feature vector derived from the players, as we
describe below.
From this, we can predict the class label for the clip using 
$w_k^\intercal h_t^e$,
where the weight vector corresponding to
class $k$ is denoted by $w_k$.
 We measure the squared-hinge loss as follows:
\begin{equation}
  L =   \frac{1}{2} \sum_{t=1}^T \sum_{k = 1}^K \max (0, 1 - y_k w_k^\intercal h^e_t)^2,
\end{equation} 
where $y_k$ is $1$ if the video belongs to class $k$,
and is $-1$ otherwise.

\subsection{Attention models}
Unlike past attention models \cite{Bahdnau_arxiv14,Xu_arxiv15,Yao_arxiv15} we need to attend to a different set of
features at each time-step. There are two key issues to address in this
setting.

First, although we have different detections in each frame, they
can be connected across the frames through an object tracking
method. This could lead to better feature representation of the
players.

Second, player attention depends on the state of the event and needs to evolve
with the event.  For instance, during the start of a ``free-throw" it is
important to attend to the player making the shot. However, towards the end of
the event the success or failure of the shot can be judged by observing the
person in possession of the ball.

With these issues in mind, we first present our model which uses player tracks
and learns a BLSTM based representation for each player track. We then
also
present a simple tracking-free baseline model.

\noindent \textbf{Attention model with tracking.}
We first associate the detections
belonging to the same player into tracks using a standard
method. We use a KLT tracker combined with
bipartite graph matching \cite{Munkres_1957} to perform the data association.

The player tracks can now be used to incorporate context
from adjacent frames while computing their representation.
We do this through a separate BLSTM which learns a latent
representation for each player at a given time-step.
The latent representation of player $i$ in frame $t$ is
given by the hidden state
$h_{ti}^p$ of the BLSTM across the player-track:
\begin{equation}
  h_{ti}^p = \mbox{BLSTM}_{track}(h_{t-1,i}^p, h_{t+1,i}^p, p_{ti}).
\end{equation}

At every time-step we want the most relevant player at that
instant to be chosen. We achieve this by computing
$a_t$ as a convex combination of the player representations
at that time-step:
\begin{eqnarray} 
\label{eq:track}
  a_t^{track} & = & \sum_{i=1}^{N_t} \gamma_{ti}^{track} h_{ti}^p, \\ \nonumber
  \gamma_{ti}^{track} & = & \text{softmax} \left(\phi\left(h^f_t, h^p_{ti}, h^e_{t-1}\right); \tau\right),
\end{eqnarray}where $N_t$ is the number of detections in frame $t$, and $\phi()$ is a 
multi layer perceptron, similar to \cite{Bahdnau_arxiv14}. $\tau$ is the softmax temperature parameter.
This attended player representation is input to the
unidirectional event recognition LSTM in Eq.~\ref{eq:event_lstm}.
This model is illustrated in Figure~\ref{fig:model}.

\eat{
Later, we show that this method achieves better
performance at event classification and detection compared to a
tracking-free model (discussed next). However, it could result in slightly worse ``key player"
identification due to tracking errors.
}

\noindent \textbf{Attention model without tracking.}
Often, tracking people in a crowded scene can be very difficult due to
occlusions and fast movements. In such settings, it is beneficial to have a
tracking-free model. This could also allow the model to be
more flexible in switching attention between players as the event progresses.
Motivated by this, we present a model where the detections in each frame are
considered to be independent from other frames. 

We  compute the (no track) attention based player feature as shown below:
\begin{eqnarray} 
\label{eq:notrack}
  a_t^{notrack} & = & \sum_{i=1}^{N_t} \gamma_{ti}^{notrack} p_{ti},
\\ \nonumber
  \gamma_{ti}^{notrack} & = & \text{softmax} \left(\phi\left(h^f_t, p_{ti}, h^e_{t-1}\right); \tau\right),
\end{eqnarray}

Note that this is similar to the tracking based attention equations except for
the direct use of the player detection feature $p_{ti}$ in place of the
BLSTM representation $h_{ti}^p$.

\begin{table*}[!htp]
\begin{center}
\small
 \begin{tabular}{|l|c|c|c|c|c|c|c|c|c|}
  \hline
Event & IDT\cite{Wang_CVPR11} & IDT\cite{Wang_CVPR11} player & C3D \cite{Tran_arxiv14} & MIL\cite{Andrews_NIPS02} &  LRCN \cite{Donahue_arxiv14} &Only player & Avg. player & Our no track & Our track \\ \hline \hline

  3-point succ.    & 0.370 & 0.428 & 0.117 & 0.237 & 0.462   & 0.469 & 0.545 & 0.583 & \textbf{0.600} \\
  3-point fail.    & 0.501 &  0.481& 0.282 & 0.335 & 0.564   & 0.614 & 0.702 & 0.668 & \textbf{0.738} \\
  fr-throw succ. & 0.778 &  0.703& 0.642   & 0.597 & 0.876   & 0.885 & 0.809 & \textbf{0.892} & 0.882 \\
  fr-throw fail. & 0.365 &  0.623& 0.319   & 0.318 & 0.584    & \textbf{0.700} & 0.641 & 0.671 & 0.516 \\
  layup succ.      & 0.283 & 0.300 & 0.195 & 0.257 & 0.463   & 0.416 & 0.472 & 0.489 & \textbf{0.500} \\
  layup fail.      & 0.278 &0.311  & 0.185 & 0.247 & 0.386   & 0.305 & 0.388 & 0.426 & \textbf{0.445} \\
  2-point succ.    & 0.136 &  0.233 & 0.078& 0.224 & 0.257    & 0.228 & 0.255 & 0.281 & \textbf{0.341} \\
  2-point fail.    & 0.303 &  0.285 & 0.254& 0.299 & 0.378   & 0.391 & \textbf{0.473} & 0.442 & 0.471 \\
  sl. dunk succ.  & 0.197 &  0.171 & 0.047 & 0.112 & 0.285   & 0.107 & 0.186 & 0.210 & \textbf{0.291} \\
  sl. dunk fail.  & 0.004 &  0.010 & 0.004 & 0.005 & \textbf{0.027} & 0.006 & 0.010 & 0.006 & 0.004 \\
  steal            & 0.555 &  0.473& 0.303 & 0.843 & 0.876 &  0.843 & \textbf{0.894} & 0.886 & 0.893 \\ \hline \hline
Mean             & 0.343 &  0.365 & 0.221  & 0.316 & 0.469 & 0.452 & 0.489 & 0.505 & \textbf{0.516} \\ \hline
  \end{tabular}
\end{center}
  \caption{Mean average precision for event {\em classification} given
    isolated clips.}
  \label{tab:event_class}
  \label{tab:class_res}
\end{table*}

\begin{table*}[ht!]
\begin{center}
\small
 \begin{tabular}{|l|c|c|c|c|c|c|c|c|}
  \hline
Event & IDT\cite{Wang_CVPR11} & IDT player\cite{Wang_CVPR11} & C3D \cite{Tran_arxiv14} & LRCN \cite{Donahue_arxiv14} & Only player & Avg. player & Attn no track & Attn track \\ \hline \hline
3-point succ.    & 0.194  & 0.203 &  0.123 & 0.230 & 0.251 & \textbf{0.268} & 0.263 & 0.239 \\
3-point fail.    & 0.393  & 0.376 &  0.311 & 0.505 & 0.526 & 0.521 & 0.556 & \textbf{0.600} \\
free-throw succ. & 0.585  & 0.621 &  0.542 & 0.741 & 0.777 & \textbf{0.811} & 0.788 & 0.810 \\
free-throw fail. & 0.231  & 0.277 &  0.458 & 0.434 & \textbf{0.470} & 0.444 & 0.468 & 0.405 \\
layup succ.      & 0.258  & 0.290 &  0.175 & 0.492 & 0.402 & 0.489 & 0.494 & \textbf{0.512} \\
layup fail.      & 0.141  & 0.200 &  0.151 & 0.187 & 0.142 & 0.139 & 0.207 & \textbf{0.208} \\
2-point succ.    & 0.161  & 0.170 &  0.126 & 0.352 & 0.371 & \textbf{0.417} & 0.366 & 0.400 \\
2-point fail.    & 0.358  & 0.339 &  0.226 & 0.544 & 0.578 & \textbf{0.684} & 0.619 & 0.674 \\
  slam dunk succ.& 0.137  & 0.275 &  0.114 & 0.428 & 0.566 & 0.457 & 0.576 & \textbf{0.555} \\
slam dunk fail.  & 0.007  & 0.006 &  0.003 & \textbf{0.122} & 0.059 & 0.009 & 0.005 & 0.045 \\
steal            & 0.242  & 0.255 &  0.187 & \textbf{0.359} & 0.348 & 0.313 & 0.340 & 0.339 \\ \hline \hline
Mean             & 0.246  & 0.273 &  0.219 & 0.400 & 0.408 & 0.414 & 0.426 & \textbf{0.435} \\ \hline
  \end{tabular}
\end{center}
  \caption{Mean average precision for event {\em detection} given
    untrimmed videos.}
  \label{tab:detection_res}
\end{table*}

\section{Experimental evaluation}
\label{sec:experiments}

In this section, we present three sets of experiments on the NCAA basketball
dataset: 1. event classification, 2. event detection and 3. evaluation of
attention.

\subsection{Implementation details}

We used a hidden state dimension of $256$ for all the LSTM and BLSTM RNNs, an
embedding layer with ReLU non-linearity and $256$ dimensions for embedding the
player features and frame features before feeding to the RNNs.  We used $32
\times 32$ bins with spatial pyramid pooling for the player location feature.
All the event videos clips were four seconds long and subsampled to 6fps.  The
$\tau$ value was set to $0.25$ for the attention softmax weighting. We used a
batch size of $128$, and a learning rate of $0.005$ which was reduced by a factor of
$0.1$ every $10000$ iterations with RMSProp\cite{RMSProp}. The models were
trained on a cluster of $20$ GPUs for $100k$ iterations over one day.  The
hyperparameters were chosen by cross-validating on the validation set.

\subsection{Event classification}

In this section, we compare the ability of methods to classify isolated
video-clips into 11 classes.  We do not use any additional
negatives from other parts of the basketball videos.  We compare our results
against different control settings and baseline models explained below:

\begin{itemize}\denselist
  \item \emph{IDT\cite{Wang_CVPR11}} We use the publicly available implementation of dense trajectories with
  Fisher encoding.
  
  \item \emph{IDT\cite{Wang_CVPR11} player} We use IDT along with averaged features extracted from the player
  bounding boxes.

  \item \emph{C3D \cite{Tran_arxiv14}} We use the publicly available pre-trained model for feature extraction
  with an SVM classifier.

  \item \emph{LRCN \cite{Donahue_arxiv14}} We use an LRCN model with frame-level features. However, we use
    a BLSTM in place of an LSTM. We found this to improve performance. Also, we do not back-propagate
    into the CNN extracting the frame-level features to be consistent with our model.

  \item \emph{MIL \cite{Andrews_NIPS02}} We use a multi-instance learning
    method to learn bag (frame) labels from the set of player features.

\item \emph{Only player} We only use our player features from Sec.~\ref{sec:feature_extraction} in our model
  without frame-level features.
 
  \item \emph{Avg. player} We combine the player features by simple averaging, without
using  attention.

  \item \emph{Attention no track} Our model without tracks (Eq.~\ref{eq:notrack}).

  \item \emph{Attention with track} Our model with tracking (Eq.~\ref{eq:track}).
\end{itemize}

The mean average precision (mAP) for each setting is shown in Tab.~\ref{tab:class_res}. We see
that the method that uses both global information and local player information
outperforms the model only using local player information (``Only player'') and only
using global information (``LRCN'').  We also show that combining the player
information using a weighted sum (i.e., an attention model) is better than
uniform averaging (``Avg. player''), with the tracking based version of
attention slightly better than the track-free version.  Also, a standard
weakly-supervised approach such as MIL seems to be less effective than any of
our modeling variants.

The performance varies by class.
In particular, performance is much poorer (for all methods) for classes
such as ``slam dunk fail'' for which we have very little data.
However, performance is better for shot-based events like
``free-throw'', ``layups'' and ``3-pointers''where attending to
the shot making person or defenders can be useful. 
\eat{
On the other hand,
``steal'' does not benefit from attention, since it is easy to
distinguish this class from others.
which
is the only event without an attempted shot is easy to distniguish
from the others and does not benefit much from attention.
}

\subsection{Event detection}

In this section, we evaluate the ability of methods to temporally localize
events in untrimmed videos.  We use a sliding window approach, where we slide a
$4$ second window through all the basketball videos and try to classify the
window into a negative class or one of the 11 event classes. We use a stride
length of $2$ seconds.  We treat all windows which do not overlap more than $1$
second with any of the $11$ annotated events as negatives. We use the same setting
for training, test and validation.  This leads to $90200$ negative examples
across all the videos.  We compare with the same baselines as before. However,
we were unable to train the MIL model due to computational limitations.

The detection results are presented in Tab.~\ref{tab:detection_res}.  We see
that, as before, the attention models beat previous state of the art methods.
Not surprisingly, all methods are slightly worse at temporal localization than
for classifying isolated clips.  We also note a significant
difference in classification and detection performance for ``steal" in all
methods.  This can be explained by the large number of negative instances
introduced in the detection setting. These negatives often correspond to
players passing the ball to each other. The ``steal" event is quite similar to
a ``pass" except that the ball is passed to a player of the opposing team. This
makes the ``steal" detection task considerably more challenging.

\subsection{Analyzing attention}

We have seen above that attention can improve the performance of the
model at tasks such as classification and detection. 
Now, we evaluate how accurate the attention models are at
identifying the key players. (Note that the models were never
explicitly trained to identify key players).

\begin{figure*}[t!]
\begin{center}
   \includegraphics[width=0.90\linewidth]{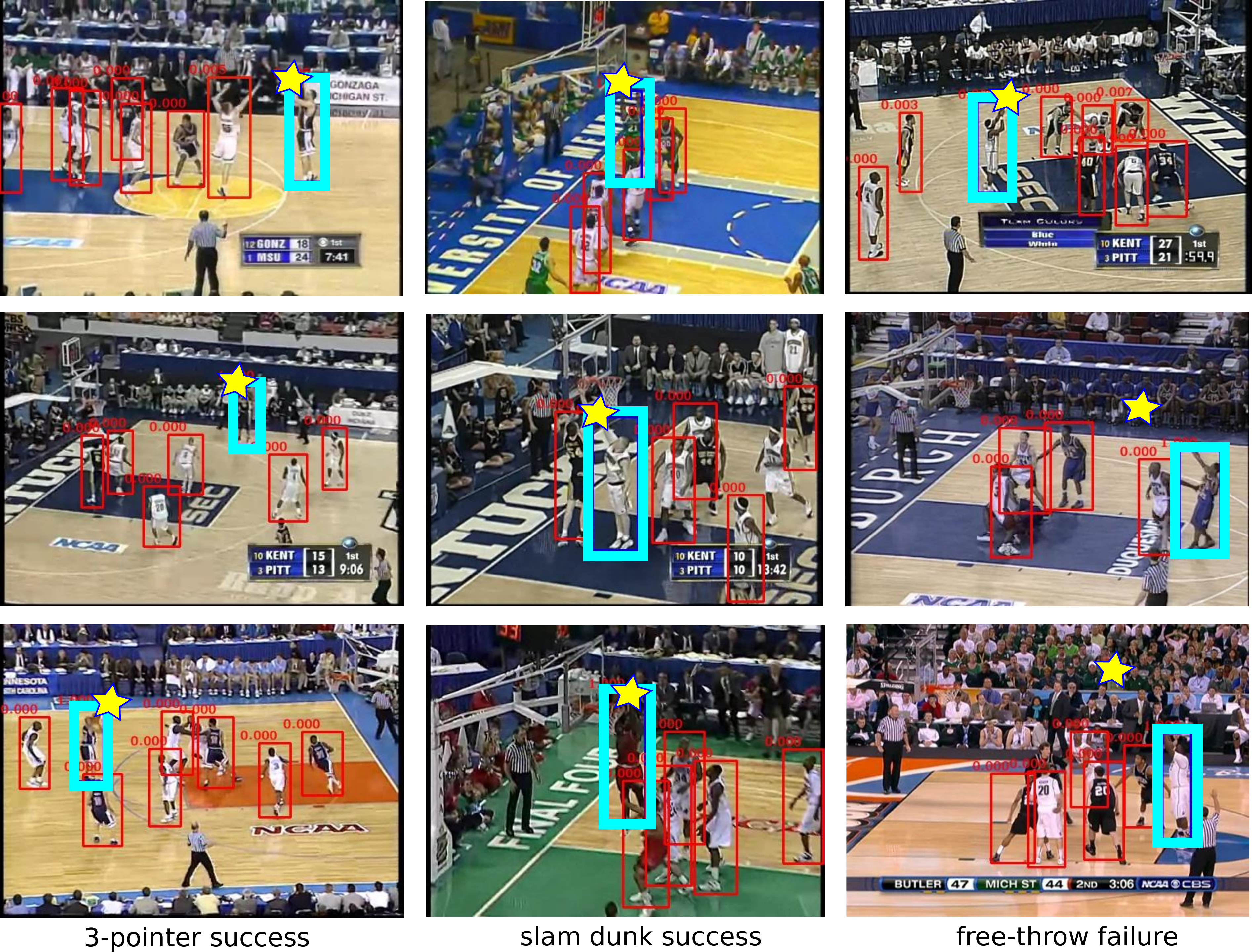}
\end{center}
   \caption{We highlight (in cyan) the ``attended" player at the beginning of different events.
     The position of the ball in each frame is shown in yellow.
   Each column shows a different event. In these videos, the model attends
 to the person making the shot at the beginning of the event.}
\label{fig:visual_attention}
\end{figure*}

\eat{
While attention to specific players improves event detection,
the attention scores themselves carry valuable information.
We observed that the attention scores represented a 
consistent meaning across multiple videos in our dataset.
More concretely, our model often ``attends" to the person shooting the
ball at the beginning of an event. We can see several visual examples
in Fig.~\ref{fig:visual_attention}, where the person shooting
the ball is highlighted by attention.}

\begin{figure*}[t!]
\begin{center}
  \vspace{-7mm}
  \includegraphics[width=0.9\linewidth]{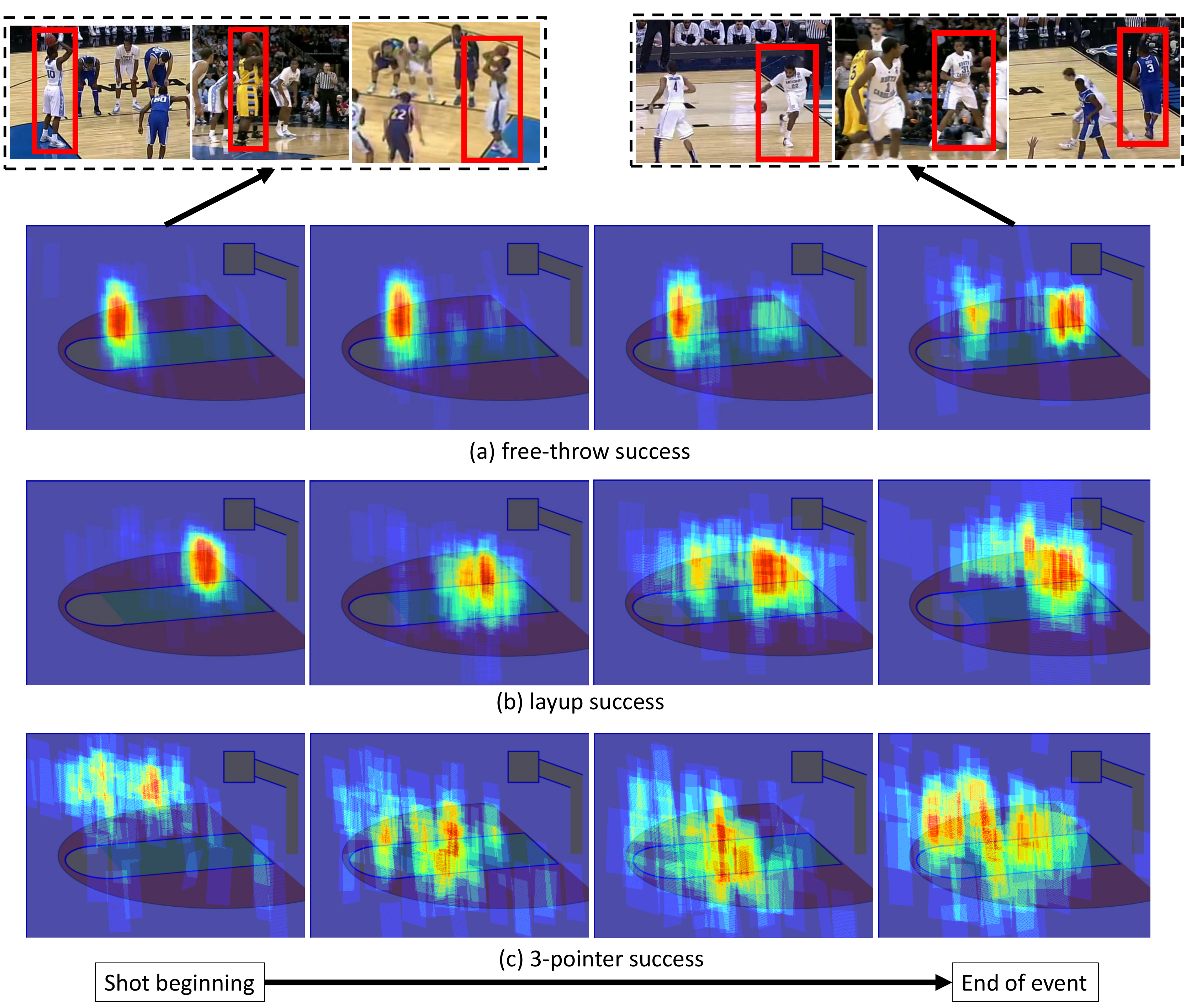}
\end{center}
  \vspace{-4mm}
   \caption{We visualize the distribution of attention over different positions of
   a basketball court as the event progresses. This is shown for 3 different events.
   These heatmaps were obtained by first transforming all videos
 to a canonical view of the court (shown in the background of each heatmap). The top row shows
 the sample frames which contributed to the ``free-throw" success heatmaps. It is interesting
 to note that the model focuses on the location of the shooter at the beginning of an
 event and later the attention disperses to other locations.}
  \vspace{-4mm}
\label{fig:att_heatmap}
\end{figure*}

To evaluate the attention models, we  labeled the player who was
closest (in image space) to the ball as the ``shooter''.
(The ball location is annotated in 850 test clips.)
We used these annotations to evaluate if our ``attention" scores
were capable of classifying the ``shooter" correctly in these frames.

\begin{table}[ht!]
\begin{center}
\small
 \begin{tabular}{|l|c|c|c|}
  \hline
Event            & Chance & Attn. with track & Attn. no track \\ \hline \hline
3-point succ.    & 0.333 & 0.445 & 0.519 \\ 
3-point fail.    & 0.334 & 0.391 & 0.545 \\ 
free-throw succ. & 0.376 & 0.416 & 0.772 \\ 
free-throw fail. & 0.346 & 0.387 & 0.685 \\  
layup succ.      & 0.386 & 0.605 & 0.627 \\ 
layup fail.      & 0.382 & 0.508 & 0.605 \\ 
2-point succ.    & 0.355 & 0.459 & 0.554 \\ 
2-point fail.    & 0.346 & 0.475 & 0.542 \\ 
slam dunk succ.  & 0.413 & 0.347 & 0.686 \\ 
slam dunk fail.  & 0.499 & 0.349 & 0.645 \\ \hline \hline  
Mean             & 0.377 & 0.438 & 0.618 \\ \hline
  \end{tabular}
\end{center}
  \caption{Mean average precision for attention evaluation.}
  \vspace{-4mm}
  \label{tab:attention_res}
\end{table}

The mean AP for this ``shooter"  classification is listed in
Tab.~\ref{tab:attention_res}.  The results show that the track-free attention
model is quite consistent in picking the shooter for several classes like
``free-throw succ./fail", ``layup succ./fail." and ``slam dunk succ.". This is
a very promising result which shows that attention on player detections alone
is capable of localizing the player making the shot. This could be a useful cue
for providing more detailed event descriptions including the identity and
position of the shooter as well.

\begin{figure}[ht!]
\begin{center}
   \includegraphics[width=0.9\linewidth]{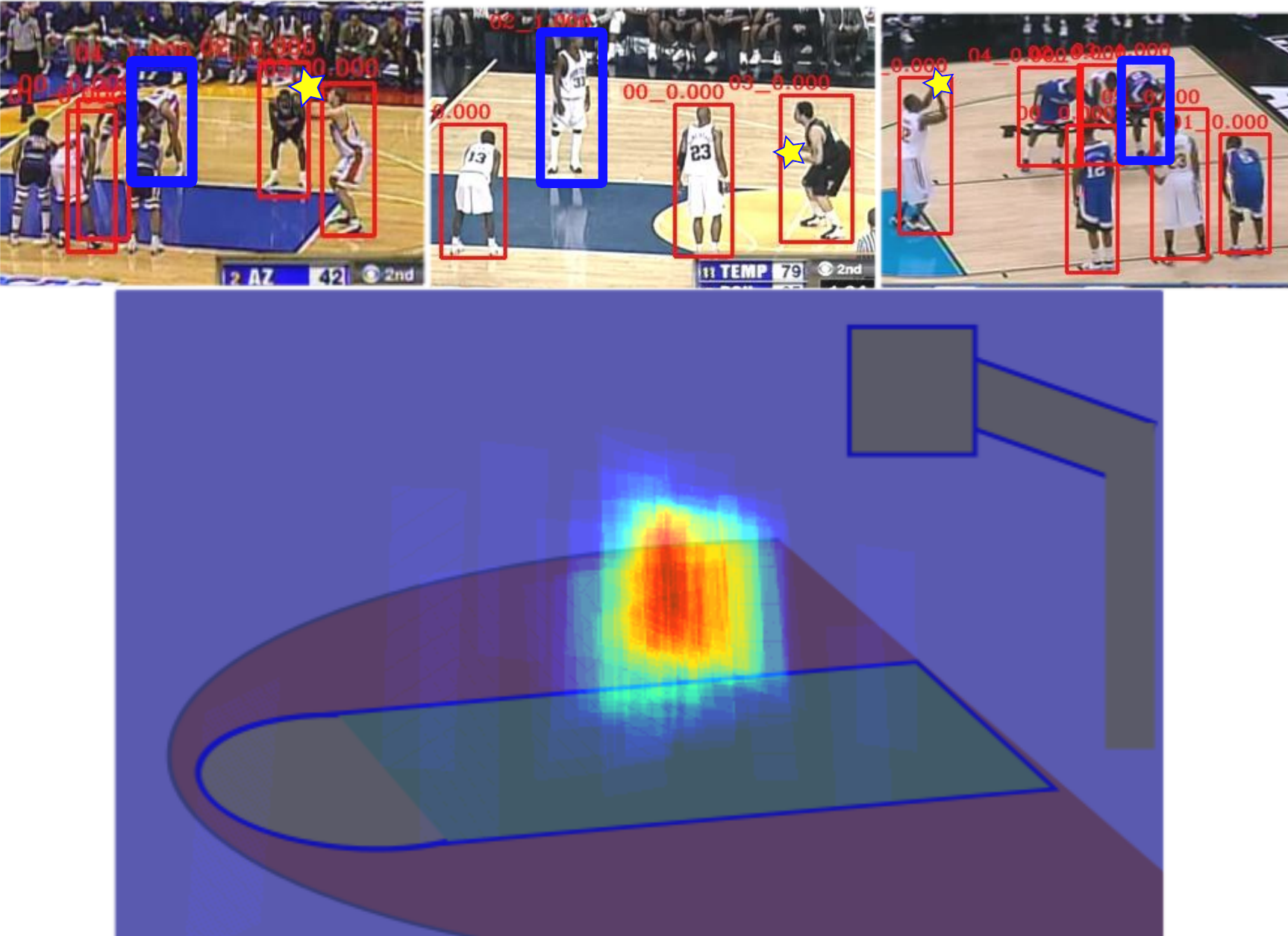}
\end{center}
  \vspace{-4mm}
\caption{The distribution of attention for our model with tracking,
     at the beginning of ``free-throw success". Unlike
     Fig.~\ref{fig:att_heatmap}, the attention is concentrated at a specific
     defender's position. Free-throws have a distinctive defense formation, and
   observing the defenders can be helpful as shown in the sample images in
 the top row.} \eat{While the model does not choose the shooter in this case, it
 consistently attends to another relevant player.}
  \vspace{-4mm}
\label{fig:visual_attention_trackspec}
\end{figure}

In addition to the above quantitative evaluation, we wanted to visualize the
attention masks visually.  Figure~\ref{fig:visual_attention} shows 
sample videos.  In order to make results comparable across frames, we annotated
5 points on the court and aligned all the attended boxes for an event to one
canonical image.  Fig.~\ref{fig:att_heatmap} shows a heatmap  visualizing the
spatial distributions of the attended players with respect to the court. It is
interesting to note that our model consistently focuses under the basket for a
layup, at the free-throw line for free-throws and outside the 3-point ring for
3-pointers.

Another interesting observation
is that the attention scores for the tracking based model are less selective in
focusing on the shooter.  We observed that the tracking model is often
reluctant to switch attention between frames and focuses on a single
player throughout the event. This biases the model towards players who are
present throughout the video. For instance, in free-throws
(Fig.~\ref{fig:visual_attention_trackspec}) the model always
attends to the defender at a specific position, who is visible throughout the
entire event unlike the shooter.

\vspace{-2mm}

\section{Conclusion}
We have introduced a new attention based model for event classification and
detection in multi-person videos. Apart from recognizing the event, our model
can identify the key people responsible for the event without being explicitly
trained with such annotations. Our method can generalize to any
multi-person setting. However, for the purpose of this paper we introduced a
new dataset of basketball videos with dense event annotations and compared our
performance with state-of-the-art methods on this dataset.  We also evaluated
the ability of our model to recognize the ``shooter" in the events with
visualizations of the spatial locations attended by our model.


\end{document}